\documentclass[letterpaper, 10 pt, conference]{ieeeconf}  

\IEEEoverridecommandlockouts                              

\usepackage{soul}
\usepackage{url}
\usepackage[hidelinks]{hyperref}
\usepackage[utf8]{inputenc}
\usepackage[small]{caption}
\usepackage{graphicx}
\usepackage{amsmath}
\usepackage{booktabs}
\usepackage{algorithm}
\usepackage[table,dvipsnames]{xcolor}  
\usepackage{amssymb}
\usepackage{mathrsfs}
\usepackage{mathtools}  
\usepackage{mathtools}

 \usepackage[normalem]{ulem}
 \usepackage{cancel}

 \usepackage{hhline}
 \usepackage{tabularx,colortbl}

 \usepackage{comment}

 \usepackage{multirow}
 \usepackage{booktabs}
 \usepackage[noend]{algpseudocode}

\title{Risk-Aware Reasoning for Autonomous Vehicles}
\author{
Majid Khonji, 
Jorge Dias, and Lakmal Seneviratne
  \thanks{Majid Khonji, Jorge Dias, and Lakmal Seneviratne are with KU Center for Autonomous Robotic Systems, Khalifa University, Abu Dhabi, UAE (email
  {\tt \small \{majid.khonji, jorge.dias, lakmal.seneviratne\}@ku.ac.ae).}}
}

\begin{document}

\maketitle
\begin{abstract}
A significant barrier to deploying autonomous vehicles (AVs) on a massive scale is safety assurance. Several technical challenges arise due to the uncertain environment in which AVs operate such as road and weather conditions, errors in perception and sensory data, and also model inaccuracy. In this paper, we propose a system architecture for risk-aware AVs capable of reasoning about uncertainty and deliberately bounding the risk of collision below a given threshold. We discuss key challenges in the area, highlight recent research developments, and propose future research directions in three subsystems. First, a perception subsystem that detects objects within a scene while quantifying the uncertainty that arises from different sensing and communication modalities. Second, an intention recognition subsystem that predicts the driving-style and the intention of agent vehicles (and pedestrians). Third, a planning subsystem that takes into account the uncertainty, from perception and intention recognition subsystems, and propagates all the way to control policies that explicitly bound the risk of collision. We believe that such a white-box approach is crucial for future adoption of AVs on a large scale.

\end{abstract}

\vspace{-3pt}
\section{Introduction}
Over the past hundred years,  innovation within the automotive industry has created more efficient, affordable, and safer vehicles, but progress has been incremental so far. The industry now is on the verge of a substantial change due to the advancements in Artificial Intelligence (AI) and Autonomous Vehicle (AV) sensing technologies. These advancements offer the possibility of significant benefits to society, saving lives, and reducing congestion and pollution. Despite the progress, a significant barrier to large scale deployment is safety assurance. 
Most technical challenges are due to the uncertain environment in which AVs operate such as road and weather conditions, errors in perception and sensory input data, and uncertainty in the behavior of the pedestrians and agent vehicles. A robust AV control algorithm should account for different sources of uncertainty and generate control policies that are quantifiably safe. In addition, algorithms that respect precise safety measures can  assist policymakers  addressing legislative issues related to AVs, such as insurance policies and ultimately convince the public for a wide deployment of AVs.

One of the most prevalent measures for AV safety is the number of crashes per million miles \cite{kalra2016driving}. Although such a measure provides some estimate on overall safety performance in a particular environment, it fails to capture unique differences and the richness of individual scenarios. As AVs become more prevalent, the reasoning behind individual events becomes of critical importance as the public would require transparency and explainable AI. Recent AV fatal crashes raise further debates among scholars and pioneers in the industry concerning how an autonomous vehicle should act when human safety is at risk. On a more philosophical level, a study \cite{bonnefon2016social} sheds light on the major challenges of understanding societal expectations about the principles that should guide the decision making in life-critical situations. As an illustrative example, suppose a self-driving vehicle, experiencing a partial system failure, forced into an ultimatum choice between running over pedestrians or sacrificing itself and its passenger to save them. What should be the reasoning behind such a situation, and more fundamentally, what should be the moral choice? Despite the profound philosophical dilemma and the impact on the public perception of AI as a whole and the regulatory aspects for AVs in particular, the current state-of-the-art of the technological stack of AVs does not explicitly capture and propagate uncertainty sufficiently well throughout decision processes in order to accurately assess these edge scenarios.

In this work, we discuss algorithmic pipeline and a technical stack for AVs to capture and propagate uncertainty from the environment throughout perception, prediction, planning, and control. An AV has to be able to plan and optimize trajectories from its current location to a goal while avoiding static and dynamic (moving) obstacles, while meeting deadlines and efficiency constraints. The risk of collision should be bounded by a given safety threshold that meets governmental regulations, while meeting deadlines should meet a quality of service threshold. 

To expand AV perception range, we consider the Vehicular Ad-Hoc Network (VANET) communication model. Vehicle-to-Vehicle (V2V), Vehicle-to-Infrastructure (V2I), and more recently Vehicle-to-Everything (V2X), are technologies that enable vehicles to exchange safety and mobility information between each other and with the surrounding agents, including pedestrians with smart phones and smart wearables. Vehicles can collect information en route, such as road conditions and position estimates of static and dynamic objects, and can use this information to continuously predict actions performed by other vehicles and infrastructure.  V2V messages would have a range of approximately 300 meters, which exceeds the capabilities of systems with cameras, ultrasonic sensors, and LIDAR, allowing greater capability and time to warn vehicles. 

In this work, we propose a system architecture (Sec.~\ref{sec:arch}) and discuss key challenges in quantifying uncertainty at different levels of abstractions: scene representation (Sec.~\ref{sec:percept}), intention recognition (Sec.~\ref{sec:intent}), risk-bounded planning (Sec.~\ref{sec:plan}), and control (Sec.~\ref{sec:motion}). We highlight current state-of-the-art, and propose research directions at each level.
\vspace{-5pt}
\section{System Architecture}\label{sec:arch}
\begin{figure}[!ht]\hspace*{-20pt}
    \centering
    \includegraphics[scale=.5]{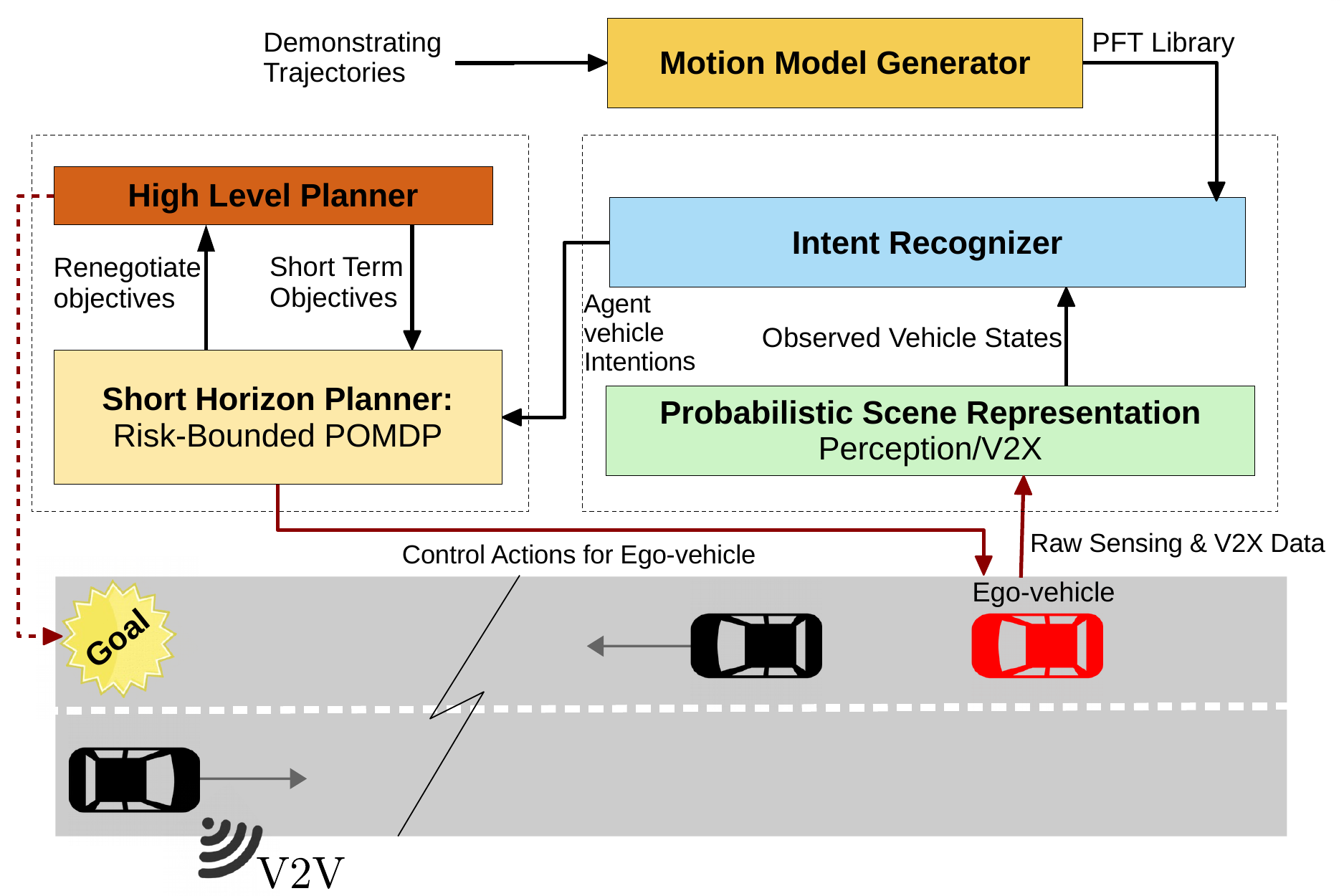}\vspace{-10pt}
    \caption{Risk-aware AV stack.}\vspace{-10pt}
    \label{fig:arch}
\end{figure}
In the following, we present the architecture of a risk-aware AV stack with six technical objectives in mind:
\begin{itemize}
\item A probabilistic perception and object representation system that takes into consideration uncertainty that arises from hardware modalities and sensor fusion. The system will capture uncertainty in object classification, bounding geometries, and temporal inconsistencies under diverse conditions. 
\item Leverage the communication network to gain knowledge of the surrounding agents (vehicles and pedestrians) that are beyond line-of-sight, and then improve upon scene representation.
\item An intention recognition system that takes into account all dynamic objects (vehicles and pedestrians), from perception and V2X communication, and estimates a distribution over potential future trajectories.
\item Generalize upon recently developed risk-aware optimization algorithms \cite{ijcai2019-775, santana2016rao}, in order to ensure that movements are safe.
\item On a higher level,  propose goal-directed autonomous planners that strive to meet the passenger goals and preferences, and help the passengers to think through adjustments to their goals, when they can't be safely met.
\item To ensure that decisions are made in a timely manner, design polynomial-time approximation algorithms that offer formal bounds on sub-optimality, and which produce near-optimal results. 
\end{itemize}

In addition, by specifying the probability that a plan is executed successfully, the system operator or policymaker can set the desired level of conservatism in the plan in a meaningful manner and can trade conservatism against performance. 
 Fig.~\ref{fig:arch} shows the interaction between key components of the system as we illustrate throughout the paper.

\vspace{-10pt}
\section{Probabilistic Scene Representation}\label{sec:percept}



 Scene understanding is research topic with strong impact on technologies for autonomous vehicles. Most of the efforts have been concentrated on understanding the scenes surrounding the ego-vehicle (autonomous vehicle itself). This is composed by sensor data processing pipeline that includes different stages such as low-level vision tasks, detection, tracking and segmentation of the surrounding traffic environment --e.g., pedestrian, cyclists and vehicles. However, for an autonomous vehicle, these low-level vision tasks are insufficient to comprehensive scene understanding. It is necessary to include reasoning about the past and the present of the scene participants.  This paper intends to guide future research on interpretation of traffic scene in autonomous driving from a probabilistic event reasoning  perspective. 
\vspace{-5pt}
\subsection{Probabilistic Context Layout for Driving}
Scene representation includes context representations that include spatially geometrical relationships \cite{Landsiedel2017} among different traffic elements with certain semantic labels. It is different from the semantic segmentation frameworks \cite{Levinkov2013}, \cite{Zhang2016}, because the context representation does not only contain the static components of traffic scene (typical technique for this aspect is simultaneous localization and mapping (SLAM)), such as road, the type of traffic lanes, traffic direction, and participant orientation, but also consists of several kinds of dynamic elements, e.g., motion correlation of participants. The study \cite{Janai2017},\cite{Zhao2017} has given a detailed review on semantic segmentation, taking the traffic geometry inferring into consideration.

A key aspect of context representation is to extract salient features from a large set of sensor data. For that purpose, it is necessary to establish a saliency mechanism, that is a critical region extraction and information simplification technique that is widely used for attractive region selection in images. Over the past few decades, saliency has been generally formulated as bottom-up and top-down modes. Bottom-up modes \cite{Moussaid2010},  \cite{Wang2013} are fast, data-driven, pre-attentive and task-independent. Top-down approaches \cite{He2016}, \cite{Yang2017}, \cite{Pan2016}, \cite{Xia2017} often entail supervised learning with pre-collected task labels by a large set of training examples and are task-oriented and vary in different environments.

A recent work \cite{senanayake2018automorphing} presents a fast algorithm that obtains a probabilistic occupancy model for dynamic obstacles in the scene with few sparse LIDAR measurements. Typically the occupancy states exhibit highly nonlinear patterns that cannot be captured with a simple linear classification model. Therefore, deep learning models and kernel-based models can be considered as potential candidates. However, these approaches require either a massive amount of data or a high number of hyper-parameters to tune. 
A promising future direction is to extend this approach to account for different object classes (rather than occupancy map) and other sensors as well such as cameras.

\vspace{-5pt}
\subsection{Beyond Line-of-sight}

\begin{figure}[!ht]
\centering
\includegraphics[scale=.45]{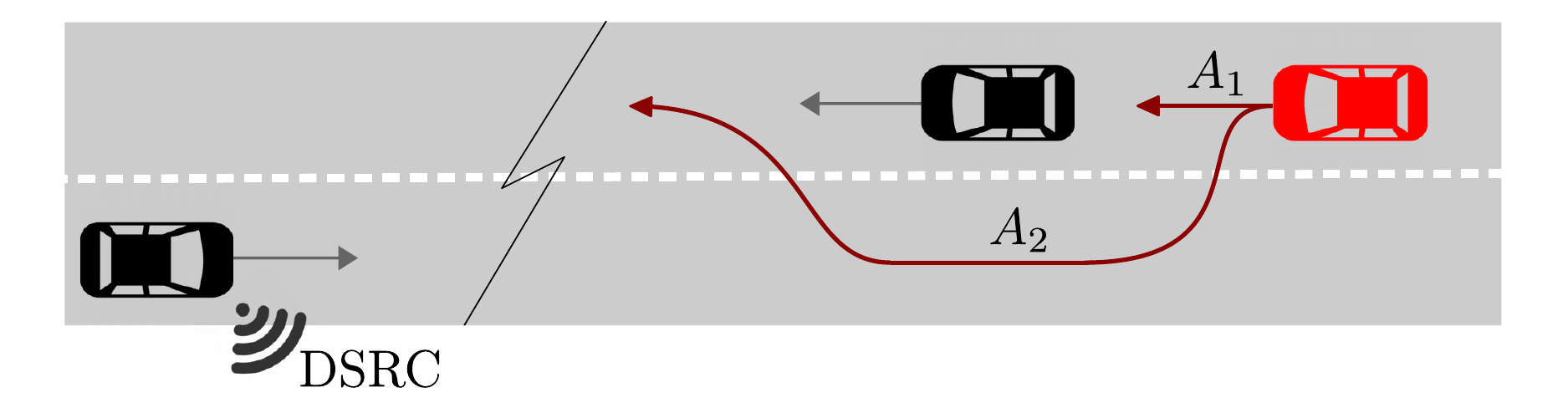}
\vspace{-10pt}
\caption{V2V communication.}\vspace{-10pt}
\label{fig:v2v}
\end{figure}

Any sensing modality has blind spots. For objects that lie beyond-line-of-sight, one can consider a communication network to improve upon the scene representation. This can be critical in certain edge scenarios. For example, in Fig.~\ref{fig:v2v}, the ego-vehicle (red) has two options: either maintain speed or overtake the vehicle ahead.  Suppose that another agent vehicle is approaching from a distance that is not detected by onboard sensors of the ego-vehicle. In this scenario,  both the speed and location of the distant vehicle might not be accurately estimated, therefore maneuver $A_2$ leading to a collision. 
 
There has been substantial progress for the standardization of vehicle-to-everything/V2X (V2V/V2I/V2P) communication protocols. The major V2X standards are known as DSRC (Dedicated Short-Range Communications) \cite{hartenstein2010vanet} as well as 5G \cite{GanteFalcao2019}. The introduction of 5G's millimeter-wave transmissions brings a new paradigm to wireless communications. Depending on the application, 5G positioning can also enhance tracking techniques, which leverage short-term historical data (local signatures and key features).
Uncertainty can be captured by probabilistic models (e.g., Gaussian) through sampling temporal inconsistencies in historical data streams such as localization data, and parameter tuning.
\vspace{-5pt}
\section{Intention Recognition}\label{sec:intent}



This subsystem involves prediction and machine learning tasks to reliably estimate the future trajectories of uncontrollable agents in the scene, including pedestrians and other agent vehicles. Many existing trajectory prediction algorithms \cite{houenou2013vehicle,woo2017lane} obtain deterministic results quite efficiently. However, these approaches fail to capture the uncertain nature of human actions. Probabilistic predictions are beneficial in many safety-critical tasks such as collision checking and risk-aware motion planning. They can express both the intrinsically uncertain prediction task at hand (human nature) and reasoning about the limitations of the prediction method (knowing when an estimate could be wrong \cite{kendall2017uncertainties}). To incorporate uncertainties into prediction results, data-driven approaches can learn common characteristics from datasets of demonstrated trajectories \cite{vasquez2009growing, wiest2012probabilistic}. These methods often express uni-modal predictions, which may not perform well in sophisticated urban scenarios where the driver can choose among multiple actions. 
A recent work \cite{huang2019uncertainty} presents a hybrid approach using a variational neural network that predicts future driver trajectory distributions for the ego-vehicle based on multiple sensors in urban scenarios. The work can be extended in future to predict trajectories for agent-vehicles using V2V  data streams, if available.

We propose a simple intent recognition that is divided into two steps. First we continuously record high-level maneuvers of surrounding vehicles (both off-line and online). Examples of such maneuvers are merge left, merge right, accelerate all at different velocities and variations and so on. Each of these maneuvers comprises of a set of collected trajectories. Due to the uncertainties in the motions of human-driven vehicles, we learn a compact motion representation called Probabilistic Flow Tube (PFT) \cite{dong2012learning} from demonstrating trajectories to capture human-like driver styles and uncertainties for each maneuver. A library of pre-learned PFTs can be used to estimate the current maneuver as well as predict the probabilistic motion of each agent vehicle using a Bayesian approach.

\vspace{-10pt}
\section{Risk-bounded Planning}\label{sec:plan}
Deterministic optimization approaches have been well developed and widely used in several disciplines and industries, in order to optimize processes both off-line and on-line. In this work, we characterize uncertainty in a probabilistic manner and find the optimal sequence of ego-vehicle trajectory control, subject to the constraint that the probability of failure must be below a certain threshold. Such constraint is known as a chance constraint. In many applications, the probabilistic approach to uncertainty modeling has a number of advantages over a deterministic approach. For instance, disturbances such as vehicle wheel slip can be represented using a stochastic model. When using a Kalman Filter for enhancing localization, the location estimate is provided as a probabilistic distribution. In addition, by specifying the probability that a plan is executed successfully, the system operator or policymaker can set the desired level of conservatism in the plan in a meaningful manner and can trade conservatism against performance. Therefore, robustness is achieved by designing solutions that guarantee feasibility as long as disturbances do not exceed these bounds. Furthermore, if the passenger goals cannot be safely achieved, then the chance constraints can be analyzed to pinpoint the sources of risk, and the user goals can be adjusted, based on their preferences, in order to restore safety.

Reasoning under uncertainty has several challenges. The optimization problem of trajectory optimization is non-convex, due to discrete choices and the presence of obstacles in the feasible space. 
One approach to tackle the challenges is by introducing multiple layers of abstractions. Instead of solving high-level problems (e.g., route planning) and low-level problems (e.g., steering wheel angle, acceleration, and brake commands) in a single shot, one can decouple them into sub-problems. We achieve such hierarchy through a high-level planner, short-horizon planner, and precomputed and learned maneuver trajectories as we illustrate below.

\vspace{-5pt}
\subsection{High Level Planner}
High-level planning involves route planning, applying traffic rules, and consequently setting short-term objectives (aka set points), which will be fed into Short Horizon Planner (as shown in Fig.~\ref{fig:arch}). The planner adjusts those short-term objectives when no safe solution exists. To be able to model the feasibility of an obtained plan, we  leverage Temporal Plan Networks (TPN) \cite{hofmann2017temporally}. A TPN is a graph where the nodes represent events, and the edges represent activities. In temporal planning, the ego-vehicle is presented with a series of events and must decide precisely when to schedule them. STNs with Uncertainty (STNUs) is an extension allowing to reason over
stochastic, or uncontrollable, actions and their corresponding durations \cite{ijcai2019-765}. Such formalism allows to check the feasibility of a high-level plan and prompt the user to adjust his or her intermediate goals and time constraints to output smooth intermediate plans, fed into the short horizon planner.

\vspace{-5pt}
\subsection{Short Horizon Planner}
Planning under uncertainty is a fundamental area in artificial intelligence. For the application of AV, it is crucial to plan for potential contingencies instead of planning a single trajectory into the future. This often occurs in dynamic environments where the vehicle has to react quickly (in milliseconds) to any potential event.
Partially observable Markov decision processes (POMDP)\cite{kaelbling1998planning, sondik1971optimal} provide a model for optimal planning under actuator and sensor uncertainty, where the goal is to find policies (contingency plans) that maximize (or minimize) some measure of expected utility (or cost). 

In many real-world applications, a single measure of performance is not sufficient to capture all requirements (e.g., an AV tasked to minimize commute time while keeping the distance from obstacle below a given threshold). This extension is often called constrained POMDP (C-POMDP) \cite{poupart2015approximate}. When constraints involve stochasticity (e.g., distance following a probabilistic model), the problem is modeled as chance-constrained POMDP (CC-POMDP) \cite{santana2016rao}, where we have a bound on the probability of violating constraints. 
To calculate the risk of each decision, one can leverage the probabilistic flow-tube (PFTs) concept to model a set of possible trajectories \cite{dong2012learning}. The current state-of-the-art solver of CC-POMDP is called RAO* \cite{santana2016rao}. RAO* generates a conditional plan based on action and risk models and likely possible scenarios for agent vehicles. 

\begin{figure}[!ht]
\centering
\includegraphics[scale=.23]{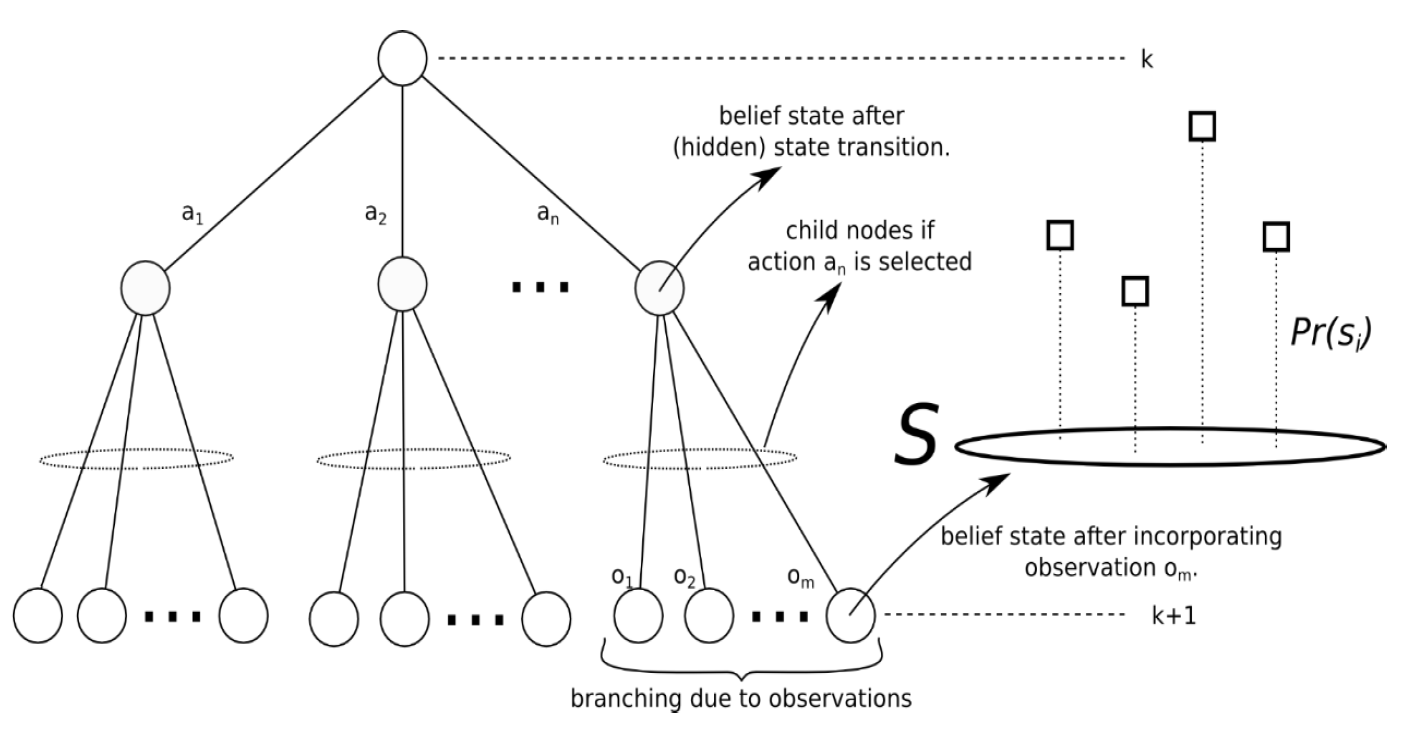}\vspace*{-5pt}
\caption{CC-POMDP Hypergraph: Nodes are the probability distributions of states (belief states) of ego vehicle. At each node, there are $n$ possible actions that can be taken by the ego vehicle. At each level, belief state is updated with respect to chosen action and observations of the environment.}\vspace{-10pt}
\label{fig:pomdp}
\end{figure}

RAO* explores from a probability distribution of vehicle states (belief state), by incrementally constructing a hypergraph, called the explicit hyper-graph shown in Fig. ~\ref{fig:pomdp}. At each node of the hyper-graph, the planner considers possible actions provided by Motion Model Generator (see Fig.~\ref{fig:arch}) and receives several possible observations. At each level, it utilizes a value heuristic to guide the search towards optimal policies. It also uses a risk heuristic to prune the search space, removing high-risk branches that violate the chance constraints. Hence, at each level, the action that maximizes expected reward and meets chance constrained is selected for the vehicle. However, one of the drawbacks of RAO* is that it does not always return optimal solutions and also does not provide any bound on the sub-optimality gap. In a recent work \cite{ijcai2019-775}, we provide an algorithm that provides guarantee on optimality (namely, a fully polynomial time approximation scheme (FPTAS)) while preserving safety constraints, all within polynomial running time. 

Recently \cite{huang2018hybrid} applied RAO* for the application of self-driving vehicles under restricted settings (e.g., known distribution of actions taken by agent-vehicles). CC-POMDP, while otherwise expressive, allow only for sequential, non-durative actions. This poses restrictions in modeling real-world planning problems. In our recent ongoing  work, we extend the framework of CC-POMDP to account for durative actions, and leverage heuristic forward search to prune the search space to improve upon the running time.

\vspace{-5pt}
\section{Motion Model Generator}\label{sec:motion}
 Based on each driving scenario, we compute a library of maneuvers. Each maneuver is associated with nominal control signals by solving a model predictive control (MPC) optimization problem \cite{huang2018hybrid}. The set of possible maneuver actions are constrained by traffic rules and vehicle dynamics and are informed by the expected evolution of the situation. Computing the actions can be accomplished through offline and online computation, and also through publicly available datasets (e.g., Berkeley DeepDrive BDD100k). 

The size of the search space of CC-POMDP, described above, is sensitive to the number of maneuver actions. To tackle this issue, we consider three different levels for abstractions. i) Micro Actions are primitive actions like Accelerate, Decelerate, Maintain. ii) Maneuver Actions are sequences of micro actions like Merge left, Merge right, iii) Macro Actions are sequences of maneuver actions such as pass the front vehicle, go straight until next intersection \cite{omidshafiei2015decentralized}. 

To calculate the risk of collision, we leverage PFT, which represents a sequence of probabilistic reachable sets. PFTs show probabilistic future predictions for states of the vehicles under a selected action. In this context, the intersection between two, temporally aligned, PFT trajectories represents the risk of collision.
To construct PFTs, we use vehicle dynamics and also probabilistic information about uncertainties, as well as through learning from datasets. By propagating the probability distributions of uncertainties through the continuous dynamics of the vehicle, we  construct probability distributions for the locations of the vehicle over a finite planning horizon. 

\vspace*{-10pt}
\section{Conclusion}
In this work, we proposed a system architecture for risk-aware AVs that can deliberately bound the risk of collision below a given threshold, defined by the policymaker. We presented the related work, discussed key challenges, and proposed research directions in three key subsystems: perception, intention recognition, and risk-aware planning. We believe that our white-box approach is crucial for a better understanding of AV decision making and ultimately for future adoption of AVs on a large scale.

\bibliographystyle{IEEEtran}
\bibliography{reference}

\begin{thebibliography}{10}
\providecommand{\url}[1]{#1}
\csname url@rmstyle\endcsname
\providecommand{\newblock}{\relax}
\providecommand{\bibinfo}[2]{#2}
\providecommand\BIBentrySTDinterwordspacing{\spaceskip=0pt\relax}
\providecommand\BIBentryALTinterwordstretchfactor{4}
\providecommand\BIBentryALTinterwordspacing{\spaceskip=\fontdimen2\font plus
\BIBentryALTinterwordstretchfactor\fontdimen3\font minus
  \fontdimen4\font\relax}
\providecommand\BIBforeignlanguage[2]{{%
\expandafter\ifx\csname l@#1\endcsname\relax
\typeout{** WARNING: IEEEtran.bst: No hyphenation pattern has been}%
\typeout{** loaded for the language `#1'. Using the pattern for}%
\typeout{** the default language instead.}%
\else
\language=\csname l@#1\endcsname
\fi
#2}}

\bibitem{kalra2016driving}
N.~Kalra and S.~M. Paddock, ``Driving to safety: How many miles of driving
  would it take to demonstrate autonomous vehicle reliability?''
  \emph{Transportation Research Part A: Policy and Practice}, vol.~94, pp.
  182--193, 2016.

\bibitem{bonnefon2016social}
J.-F. Bonnefon, A.~Shariff, and I.~Rahwan, ``The social dilemma of autonomous
  vehicles,'' \emph{Science}, vol. 352, no. 6293, pp. 1573--1576, 2016.

\bibitem{ijcai2019-775}
\BIBentryALTinterwordspacing
M.~Khonji, A.~Jasour, and B.~Williams, ``Approximability of constant-horizon
  constrained pomdp,'' in \emph{Proceedings of the Twenty-Eighth International
  Joint Conference on Artificial Intelligence, {IJCAI-19}}.\hskip 1em plus
  0.5em minus 0.4em\relax International Joint Conferences on Artificial
  Intelligence Organization, 7 2019, pp. 5583--5590. [Online]. Available:
  \url{https://doi.org/10.24963/ijcai.2019/775}
\BIBentrySTDinterwordspacing

\bibitem{santana2016rao}
P.~Santana, S.~Thi{\'e}baux, and B.~Williams, ``Rao*: an algorithm for chance
  constrained pomdps,'' in \emph{Proc. AAAI Conference on Artificial
  Intelligence}, 2016.

\bibitem{Landsiedel2017}
C.~Landsiedel and D.~Wollherr, ``Road geometry estimation for urban semantic
  maps using open data,'' \emph{Advanced Robotics}, vol.~31, no.~5, pp.
  282--290, 2017.

\bibitem{Levinkov2013}
E.~Levinkov and M.~Fritz, ``Sequential bayesian model update under structured
  scene prior for semantic road scenes labeling,'' in \emph{Proceedings of the
  IEEE International Conference on Computer Vision}, 2013, pp. 1321--1328.

\bibitem{Zhang2016}
Z.~Zhang, S.~Fidler, and R.~Urtasun, ``Instance-level segmentation for
  autonomous driving with deep densely connected mrfs,'' in \emph{Proceedings
  of the IEEE Conference on Computer Vision and Pattern Recognition}, 2016, pp.
  669--677.

\bibitem{Janai2017}
J.~Janai, F.~G{\"u}ney, A.~Behl, and A.~Geiger, ``Computer vision for
  autonomous vehicles: Problems, datasets and state-of-the-art,'' \emph{arXiv
  preprint arXiv:1704.05519}, 2017.

\bibitem{Zhao2017}
B.~Zhao, J.~Feng, X.~Wu, and S.~Yan, ``A survey on deep learning-based
  fine-grained object classification and semantic segmentation,''
  \emph{International Journal of Automation and Computing}, vol.~14, no.~2, pp.
  119--135, 2017.

\bibitem{Moussaid2010}
J.~Zhang and S.~Sclaroff, ``Exploiting surroundedness for saliency detection: a
  boolean map approach,'' \emph{IEEE transactions on pattern analysis and
  machine intelligence}, vol.~38, no.~5, pp. 889--902, 2015.

\bibitem{Wang2013}
Q.~Wang, Y.~Yuan, P.~Yan, and X.~Li, ``Saliency detection by multiple-instance
  learning,'' \emph{IEEE transactions on cybernetics}, vol.~43, no.~2, pp.
  660--672, 2013.

\bibitem{He2016}
S.~He, R.~W. Lau, and Q.~Yang, ``Exemplar-driven top-down saliency detection
  via deep association,'' in \emph{Proceedings of the IEEE Conference on
  Computer Vision and Pattern Recognition}, 2016, pp. 5723--5732.

\bibitem{Yang2017}
J.~Yang and M.-H. Yang, ``Top-down visual saliency via joint crf and dictionary
  learning,'' \emph{IEEE transactions on pattern analysis and machine
  intelligence}, vol.~39, no.~3, pp. 576--588, 2016.

\bibitem{Pan2016}
J.~Pan, E.~Sayrol, X.~Giro-i Nieto, K.~McGuinness, and N.~E. O'Connor,
  ``Shallow and deep convolutional networks for saliency prediction,'' in
  \emph{Proceedings of the IEEE Conference on Computer Vision and Pattern
  Recognition}, 2016, pp. 598--606.

\bibitem{Xia2017}
Y.~Xia, D.~Zhang, A.~Pozdnoukhov, K.~Nakayama, K.~Zipser, and D.~Whitney,
  ``Training a network to attend like human drivers saves it from common but
  misleading loss functions,'' \emph{arXiv preprint arXiv:1711.06406}, 2017.

\bibitem{senanayake2018automorphing}
R.~Senanayake, A.~Tompkins, and F.~Ramos, ``Automorphing kernels for
  nonstationarity in mapping unstructured environments.'' in \emph{CoRL}, 2018,
  pp. 443--455.

\bibitem{hartenstein2010vanet}
H.~Hartenstein and K.~Laberteaux, \emph{VANET: vehicular applications and
  inter-networking technologies}.\hskip 1em plus 0.5em minus 0.4em\relax Wiley
  Online Library, 2010, vol.~1.

\bibitem{GanteFalcao2019}
F.~G. . S.~L. Gante, J., ``Deep learning architectures for accurate millimeter
  wave positioning in 5g,'' \emph{Neural Process Letters
  https://doi.org/10.1007/s11063-019-10073}, pp. 1--28, 2019.

\bibitem{houenou2013vehicle}
A.~Houenou, P.~Bonnifait, V.~Cherfaoui, and W.~Yao, ``Vehicle trajectory
  prediction based on motion model and maneuver recognition,'' in \emph{2013
  IEEE/RSJ international conference on intelligent robots and systems}.\hskip
  1em plus 0.5em minus 0.4em\relax IEEE, 2013, pp. 4363--4369.

\bibitem{woo2017lane}
H.~Woo, Y.~Ji, H.~Kono, Y.~Tamura, Y.~Kuroda, T.~Sugano, Y.~Yamamoto,
  A.~Yamashita, and H.~Asama, ``Lane-change detection based on
  vehicle-trajectory prediction,'' \emph{IEEE Robotics and Automation Letters},
  vol.~2, no.~2, pp. 1109--1116, 2017.

\bibitem{kendall2017uncertainties}
A.~Kendall and Y.~Gal, ``What uncertainties do we need in bayesian deep
  learning for computer vision?'' in \emph{Advances in neural information
  processing systems}, 2017, pp. 5574--5584.

\bibitem{vasquez2009growing}
D.~Vasquez, T.~Fraichard, and C.~Laugier, ``Growing hidden markov models: An
  incremental tool for learning and predicting human and vehicle motion,''
  \emph{The International Journal of Robotics Research}, vol.~28, no. 11-12,
  pp. 1486--1506, 2009.

\bibitem{wiest2012probabilistic}
J.~Wiest, M.~H{\"o}ffken, U.~Kre{\ss}el, and K.~Dietmayer, ``Probabilistic
  trajectory prediction with gaussian mixture models,'' in \emph{2012 IEEE
  Intelligent Vehicles Symposium}.\hskip 1em plus 0.5em minus 0.4em\relax IEEE,
  2012, pp. 141--146.

\bibitem{huang2019uncertainty}
X.~Huang, S.~McGill, B.~C. Williams, L.~Fletcher, and G.~Rosman,
  ``Uncertainty-aware driver trajectory prediction at urban intersections,''
  \emph{arXiv preprint arXiv:1901.05105}, 2019.

\bibitem{dong2012learning}
S.~Dong and B.~Williams, ``Learning and recognition of hybrid manipulation
  motions in variable environments using probabilistic flow tubes,''
  \emph{International Journal of Social Robotics}, vol.~4, no.~4, pp. 357--368,
  2012.

\bibitem{hofmann2017temporally}
A.~G. Hofmann and B.~C. Williams, ``Temporally and spatially flexible plan
  execution for dynamic hybrid systems,'' \emph{Artificial Intelligence}, vol.
  247, pp. 266--294, 2017.

\bibitem{ijcai2019-765}
\BIBentryALTinterwordspacing
N.~Bhargava and B.~C. Williams, ``Faster dynamic controllability checking in
  temporal networks with integer bounds,'' in \emph{Proceedings of the
  Twenty-Eighth International Joint Conference on Artificial Intelligence,
  {IJCAI-19}}.\hskip 1em plus 0.5em minus 0.4em\relax International Joint
  Conferences on Artificial Intelligence Organization, 7 2019, pp. 5509--5515.
  [Online]. Available: \url{https://doi.org/10.24963/ijcai.2019/765}
\BIBentrySTDinterwordspacing

\bibitem{kaelbling1998planning}
L.~P. Kaelbling, M.~L. Littman, and A.~R. Cassandra, ``Planning and acting in
  partially observable stochastic domains,'' \emph{Artificial intelligence},
  vol. 101, no. 1-2, pp. 99--134, 1998.

\bibitem{sondik1971optimal}
E.~J. Sondik, ``The optimal control of partially observable markov decision
  processes.'' \emph{PhD the sis, Stanford University}, 1971.

\bibitem{poupart2015approximate}
P.~Poupart, A.~Malhotra, P.~Pei, K.-E. Kim, B.~Goh, and M.~Bowling,
  ``Approximate linear programming for constrained partially observable markov
  decision processes,'' in \emph{Twenty-Ninth AAAI Conference on Artificial
  Intelligence}, 2015.

\bibitem{huang2018hybrid}
X.~Huang, A.~Jasour, M.~Deyo, A.~Hofmann, and B.~C. Williams, ``Hybrid
  risk-aware conditional planning with applications in autonomous vehicles,''
  in \emph{2018 IEEE Conference on Decision and Control (CDC)}.\hskip 1em plus
  0.5em minus 0.4em\relax IEEE, 2018, pp. 3608--3614.

\bibitem{omidshafiei2015decentralized}
S.~Omidshafiei, A.-A. Agha-Mohammadi, C.~Amato, and J.~P. How, ``Decentralized
  control of partially observable markov decision processes using belief space
  macro-actions,'' in \emph{2015 IEEE International Conference on Robotics and
  Automation (ICRA)}.\hskip 1em plus 0.5em minus 0.4em\relax IEEE, 2015, pp.
  5962--5969.

\end{thebibliography}

\end{document}